\documentclass{article}

\usepackage[numbers]{natbib}

\usepackage[preprint]{neurips_2023}

\usepackage[utf8]{inputenc} 
\usepackage[T1]{fontenc}    
\usepackage{url}            
\usepackage{booktabs}       
\usepackage{graphicx}
\usepackage{amsfonts}       
\usepackage{nicefrac}       
\usepackage{microtype}      
\usepackage{amsmath}      
\usepackage{color}
\usepackage{subcaption} 
\usepackage{multirow} 
\usepackage{makecell}
\usepackage{colortbl}
\usepackage{wasysym}
\usepackage{setspace}
\usepackage{booktabs}
\usepackage{overpic}
\usepackage{textpos}
\usepackage{subfloat}
\usepackage{tabularx}
\usepackage{algorithm}
\usepackage{times}
\usepackage{float}
\usepackage{amssymb}
\usepackage[table]{xcolor}
\usepackage{listings}
\usepackage{enumitem}
\usepackage[british, english, american]{babel}
\usepackage[pagebackref=false, breaklinks=true, letterpaper=true, colorlinks, bookmarks=false]{hyperref}

\usepackage{etoolbox}
\makeatletter
\AfterEndEnvironment{algorithm}{\let\@algcomment\relax}
\AtEndEnvironment{algorithm}{\kern2pt\hrule\relax\vskip3pt\@algcomment}
\let\@algcomment\relax
\newcommand\algcomment[1]{\def\@algcomment{\footnotesize#1}}
\renewcommand\fs@ruled{\def\@fs@cfont{\bfseries}\let\@fs@capt\floatc@ruled
  \def\@fs@pre{\hrule height.8pt depth0pt \kern2pt}%
  \def\@fs@post{}%
  \def\@fs@mid{\kern2pt\hrule\kern2pt}%
  \let\@fs@iftopcapt\iftrue}
\makeatother
\definecolor{mygray}{gray}{.92}
\definecolor{mygreen}{RGB}{93,173,85}
\makeatletter
\newcommand{\thickhline}{%
    \noalign {\ifnum 0=`}\fi \hrule height 0.8pt
    \futurelet \reserved@a \@xhline
}
\makeatother
\newcommand{\hlg}[1]{\textcolor{mygreen}{#1}}
\newcommand{\sgrouptablestyle}[2]{\setlength{\tabcolsep}{#1}\renewcommand{\arraystretch}{#2}\centering}
\newcolumntype{d}[1]{>{\raggedright\arraybackslash}p{#1pt}}
\newcolumntype{e}[1]{>{\raggedleft\arraybackslash}p{#1pt}}
\newcommand{\bbetter}[4]{
    \sgrouptablestyle{1pt}{1}
    \begin{tabular}{e{#1}d{#2}}
    {#3} &
    {\fontsize{6.5pt}{1em}\selectfont \hlg{\textbf{$\uparrow$#4}}}
    \end{tabular}
}
\definecolor{baselinecolor}{gray}{.9}

\newlength\savewidth\newcommand\shline{\noalign{\global\savewidth\arrayrulewidth
  \global\arrayrulewidth 1pt}\hline\noalign{\global\arrayrulewidth\savewidth}}
\newcommand{\tablestyle}[2]{\setlength{\tabcolsep}{#1}\renewcommand{\arraystretch}{#2}\centering\footnotesize}
\renewcommand{\paragraph}[1]{\vspace{1.25mm}\noindent\textbf{#1}}

\newcolumntype{x}[1]{>{\centering\arraybackslash}p{#1pt}}
\newcolumntype{y}[1]{>{\raggedright\arraybackslash}p{#1pt}}
\newcolumntype{z}[1]{>{\raggedleft\arraybackslash}p{#1pt}}

\newcommand{\app}{\raise.17ex\hbox{$\scriptstyle\sim$}}

\definecolor{deemph}{gray}{0.6}

\definecolor{baselinecolor}{gray}{.9}

\title{Medical supervised masked autoencoders: Crafting a better masking strategy and efficient fine-tuning schedule for medical image classification}

\author{Jiawei Mao$^{1,2,\dag}$    \quad Shujian Guo$^{1,2,\dag}$  \quad Yuanqi Chang$^{1,2}$ \quad Xuesong Yin$^{1,2,}$ {\thanks{Corresponding author. $\dag$ Equal contribution.}}    \quad Binling Nie$^{2}$   \\ 
Wenzhou lnstitute of Hangzhou Dianzi University, Wenzhou, 325038, China$^{1}$  \qquad \\
School of Media and Design, Hangzhou Dianzi University, Hangzhou, 310018, China$^{2}$ \qquad \\
{\tt\small\{jiaweima0,221330017,yuanqichang,yinxs,binlingnie\}@hdu.edu.cn }\\
}

\begin{document}

\maketitle

\begin{abstract}
  Masked autoencoders (MAEs) have displayed significant potential in the classification and semantic segmentation of medical images in the last year. 
  Due to the high similarity of human tissues, even slight changes in medical images may represent diseased tissues, necessitating fine-grained inspection to pinpoint diseased tissues. The random masking strategy of MAEs is likely to result in areas of lesions being overlooked by the model. 
  At the same time, inconsistencies between the pre-training and fine-tuning phases impede the performance and efficiency of MAE in medical image classification. 
  To address these issues, we propose a medical supervised masked autoencoder (MSMAE) in this paper. In the pre-training phase, MSMAE precisely masks medical images via the attention maps obtained from supervised training, contributing to the representation learning of human tissue in the lesion area. 
  During the fine-tuning phase, MSMAE is also driven by attention to the accurate masking of medical images. This improves the computational efficiency of the MSMAE while increasing the difficulty of fine-tuning, which indirectly improves the quality of MSMAE medical diagnosis. 
  Extensive experiments demonstrate that MSMAE achieves state-of-the-art performance in case with three official medical datasets for various diseases. 
  Meanwhile, transfer learning for MSMAE also demonstrates the great potential of our approach for medical semantic segmentation tasks. Moreover, the MSMAE accelerates the inference time in the fine-tuning phase by 11.2\% and reduces the number of floating-point operations (FLOPs) by 74.08\% compared to a traditional MAE.
\end{abstract}

\section{Introduction}
With the development of deep learning, medical image classification based on deep learning \cite{azizi2021big,wang2020focalmix,qu2022bi,kothawade2022clinical,nielsen2023self,gong2021deformable} has received more and more research attention. 
Deep learning leverages the vast amount of data available to help clinicians make more accurate and efficient diagnoses. 
This may lead to earlier detection of diseases, thus avoiding delays in treating the illnesses. At the same time, medical image classification based on deep learning is a promising solution to issues such as the shortage of medical professionals and equipment. 
Therefore, it is necessary to develop an efficient medical image classification algorithm based on deep learning.

In the past year, masked autoencoders (MAEs) \cite{he2022masked} have emerged as the powerful tools for the classification \cite{xiao2023delving} and semantic segmentation \cite{xie2022automatic,lu2023dcelanm,liu2023edmae,yan2023representation,zhou2022self} of medical images. 
They have demonstrated exceptional performance in several studies related to medical image classification. 
For example, Xiao et al. \cite{xiao2023delving} successfully adapted MAEs to medical images representation learning and solved the data starvation issue of vision transformer \cite{dosovitskiy2020image}, which enables vision transformers to achieve comparable or even better performance than state-of-the-art (SOTA) convolutional neural networks (CNNs) for medical image classification. 
In another study, Xu et al. \cite{xu2022self} utilized pre-trained MAEs on ImageNet \cite{deng2009imagenet} to classify a sparse number of computerized tomography (CT) scans via transfer learning, which overcomes the issue of privacy restrictions in medical data. 
Moreover, Wu et al. proposed FedMAE \cite{wu2022federated}, a framework and incorporates MAE into unlabeled data federated learning (FL) for diagnostic work on dermatological diseases, which achieves higher accuracy compared to other SOTA methods. 
Zhou et al. \cite{zhou2022self} also demonstrated the promising application of MAE pre-training to medical image analysis by evaluating the performance of networks with pre-trained MAEs on three different medical image tasks. 
In summary, these recent studies have demonstrated the immense potential of MAEs in medical image classification and have paved the way for further research in this field.

\begin{figure}[t]
  \centering
  \setlength{\abovecaptionskip}{2mm} 
  \setlength{\belowcaptionskip}{-6mm}
  \includegraphics[width=0.9\linewidth]{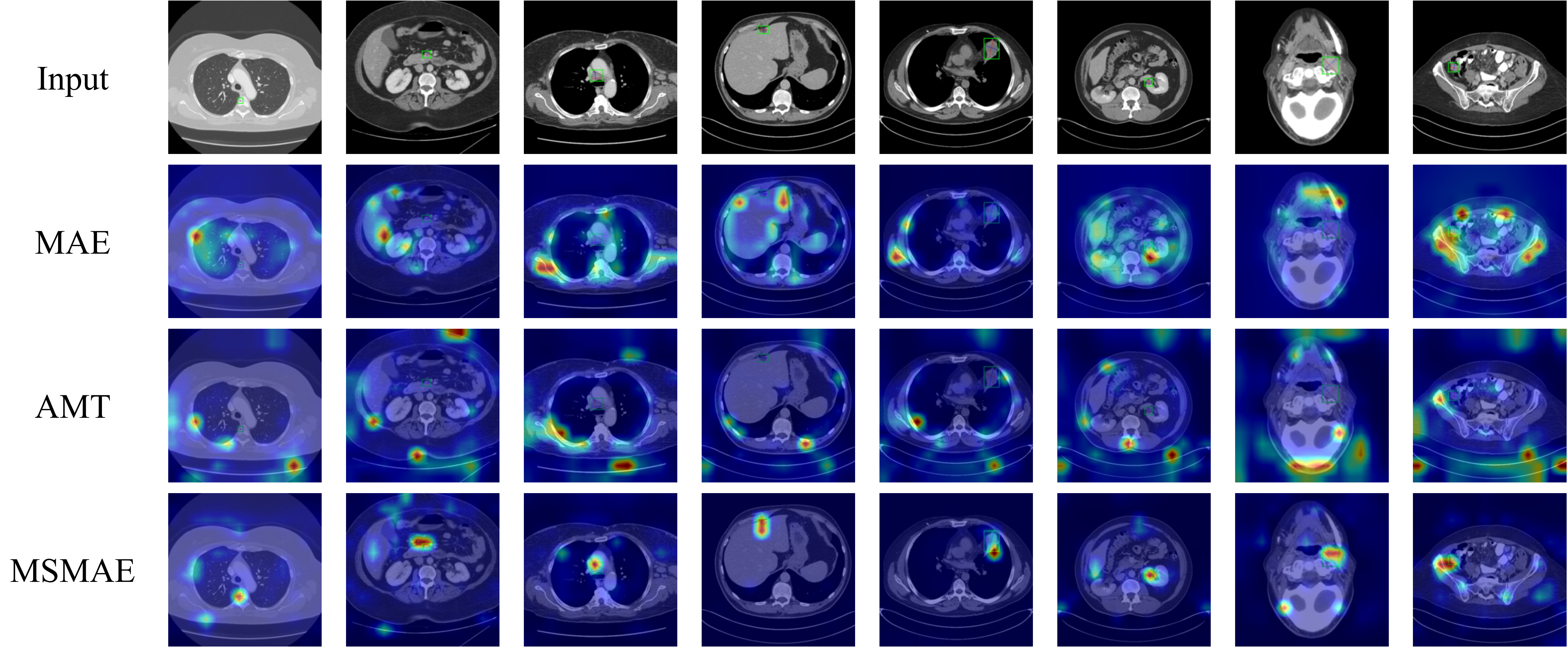}
   \caption{Attention visualizations for the self-attention of the CLS token on the heads of the last layer of the encoder after pre-training. 
   Notably, in contrast to MAE and AMT, our masking strategy enables the model to ultimately focus successfully on lesion-associated human tissue. The image inside the \textcolor{green}{green box} indicates thelesion-related human tissue. Please zoom in to see the details better.}
   \label{fig1}
\end{figure}

Although MAEs excel at in medical image classification, there are still several issues to be considered. As shown in Fig. \ref{fig1}, the diseased tissues may appear in some subtle areas of the medical image, and the random masking strategy of MAE is likely to result in the diseased tissue being overlooked by the model. 
In addition, inconsistencies arising from the pre-training and fine-tuning phases of MAE may have adverse effects on medical image classification, such as diminished performance and efficiency or even overfitting, thereby compromising the reliability and accuracy of the diagnostic results.

\begin{figure}[t]
  \centering
  \vspace{-3mm}
  \setlength{\abovecaptionskip}{2mm} 
  \setlength{\belowcaptionskip}{-3mm}
  \includegraphics[width=1\linewidth]{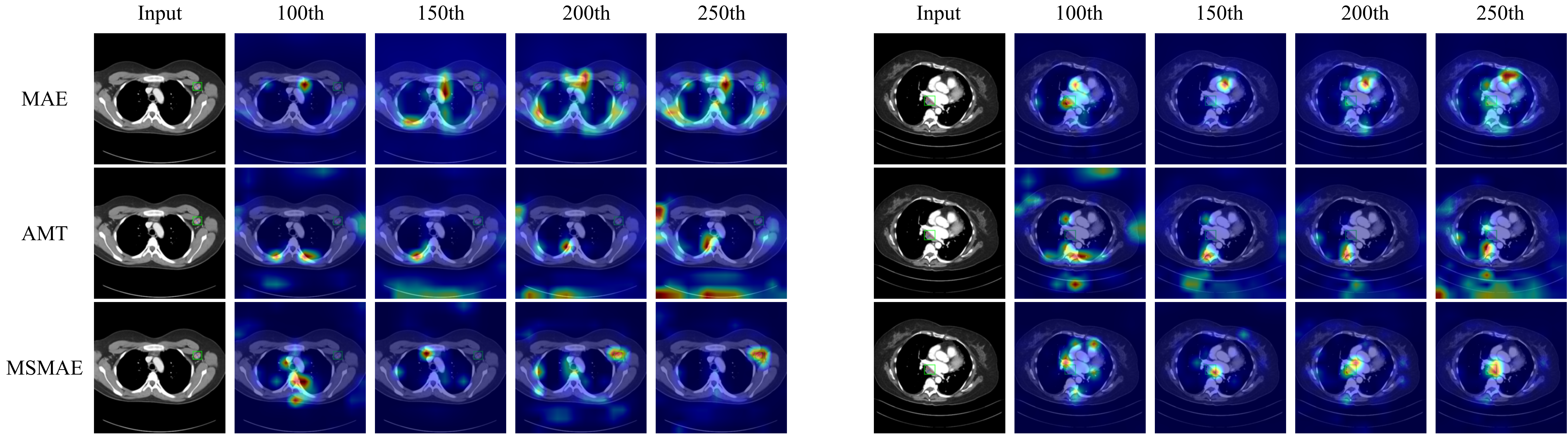}
   \caption{Results of attention visualization at various stages of the pre-training process for different algorithms. 
   It can be found that our method can pinpoint the lesion area during the pre-training process. 
   This results in MSMAE being able to accurately mask the lesion area and thus make the model more focused on the body tissue associated with the lesion.}
   \label{fig2}
\end{figure}

To address the two issues above, we propose the medical supervised masked autoencoder (MSMAE). First, in the pre-training phase, we employ supervised learning to obtain the saliency map to constrain the masking of MSMAE. We call this masking strategy a supervised attention-driven masking strategy (SAM). 
Unlike other masking strategies \cite{gui2022good,he2022masked}, SAM can precisely mask the diseased tissue, as shown in Fig. \ref{fig2}. This allows MSMAE to efficiently learn lesion-related human tissue representations during the pre-training phase. In addition, we extend this masking strategy to the fine-tuning phase of MSMAE. 
The SAM of the fine-tuning phase solves the issue of inconsistency between the pre-training and fine-tuning phases of MAEs in medical image classification. 
Furthermore, SAM improves the difficulty of MSMAE fine-tuning, thus indirectly contributing to the quality of the diagnosis. This greatly improves the computational efficiency of MSMAE during the fine-tuning phase. We demonstrate the impact of SAM on the fine-tuning phase in Fig. \ref{fig3}.

It is worth noting that our MSMAE achieves SOTA performance in several medical image classification tasks compared to various supervised and self-supervised algorithms. 
Specifically, our MSMAE outperformed the MAE by 2.87\%, 15.93\%, and 6.96\% in the Messidor-2 dataset \cite{abramoff2016improved} for the diabetic retina (DR) medical image classification task, the brain tumor MRI dataset (BTMD) \cite{amin2021brain} for the brain tumor (BT) medical image classification task, and the HAM10000 \cite{tschandl2018ham10000} dataset for the skin cancer medical image classification task, respectively. 
Meanwhile, experiments show that our MSMAE can be smoothly transferred to the medical semantic segmentation task. 
Our pre-trained MSMAE on the Breast Ultrasound Images Dataset (BUSI) \cite{al2020dataset} performs well in breast cancer segmentation task. Furthermore, compared to MAE, MSMAE reduces the inference time during fine-tuning by 11.2\% and floating-point operations (FLOPs) by 74.08\%.

In summary, the main contributions of this paper are as follows:
\begin{itemize}\vspace{-2mm}
  \item We propose a supervised attention-driven masking strategy called SAM. Compared to other masking strategies, SAM masks diseased tissue more precisely, resulting in better 	pre-training for medical image classification task. Based on such a masking strategy, we propose the medical supervised masking autoencoder (MSMAE).
  \item We also extend SAM to the fine-tuning phase of the MSMAE to maintain consistency between the pre-training and fine-tuning phases. This improves the performance of the MSMAE while substantially improving the computational efficiency of the fine-tuning phase.
  \item Extensive experiments based on three different medical image classification tasks show that the MSMAE exhibits superior performance over various supervised and self-supervised algorithms. At the same time, compared to other algorithms, MSMAE also shows some advantages in medical segmentation task. Moreover, compared with the traditional MAE, the MSMAE reduces the inference time in the fine-tuning phase by 11.2\% and the number of FLOPs by 74.08\%.
\end{itemize}

\section{Related work}
\subsection{Masked Image Modeling}

Masked image modeling has gained increasing attention in recent years as it helps the model learn features in different areas of the image and reconstructs missing images from unmasked images. 
In 2016, Pathak et al. proposed a context encoder method \cite{pathak2016context} to predict the original image by masking a rectangular area of the image. Subsequently, various novel approaches have been proposed to improve upon this method. 
Inspired by the success of BERT \cite{devlin2018bert} in natural language processing, Bao et al. proposed Beit \cite{bao2021beit}, which uses a pre-trained dVAE \cite{vahdat2018dvae} network to allow the model to directly predict discrete visual token values. 
He et al. proposed MAE \cite{he2022masked}, which employs an asymmetric encoder-decoder architecture and performs masked image modeling (MIM) tasks by encoding the image with missing pixels and then reconstructing the missing pixels. 
In contrast to MAE, MaskFeat \cite{wei2022masked} replaces direct pixel prediction with a prediction of the directional gradient histogram (HOG \cite{dalal2005histograms}) of the image. Xie et al. proposed SimMIM \cite{xie2022simmim}, 
which performs MIM tasks using medium-sized mask blocks and uses direct regression to predict raw pixel RGB values. 

\subsection{Masking Strategy}
The choice of masking strategy is an important factor in the effectiveness of Masked Image Modeling (MIM) tasks, and as such, many studies have focused on proposing new and effective masking strategies. 
AMT \cite{gui2022good} adopts an attention-based masking approach by recovering the primary object and thus reducing the model's focus on the background. Additionally, to narrow the gap between vision and language, Li et al. \cite{li2022semmae} proposed a semantic segmentation-guided masking strategy. 
On the other hand, AttMask \cite{kakogeorgiou2022hide} increases the challenge of MIM self-distillation learning by masking more discriminative cues using attention masking. 
AutoMAE \cite{chen2023improving} uses generative adversarial networks to optimize masking strategies. Wang et al. \cite{wang2023hard} proposed a novel masking strategy that autonomously masks difficult samples, improving the difficulty of the agent task. 

\section{Method}

\begin{figure}[h]
  \centering
  \vspace{-3mm}
  \setlength{\belowcaptionskip}{-1mm}
  \includegraphics[width=0.95\linewidth]{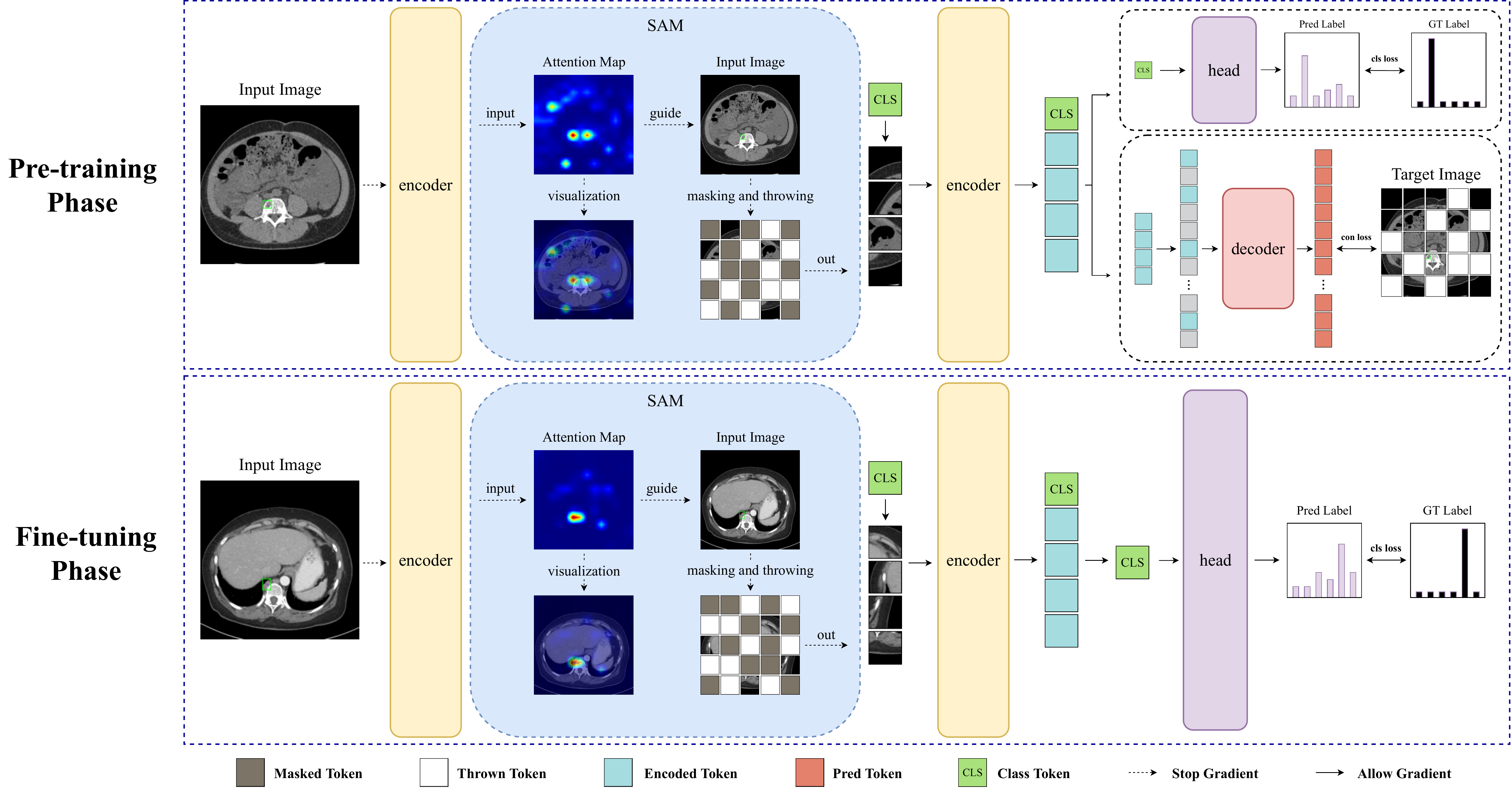}
   \caption{Detailed pipeline architecture of MSMAE.}
   \label{fig4}
   \vspace{-3mm}
\end{figure}

In this section, we elaborate on the masking strategy and fine-tuning schedule of the proposed MSMAE. 
To provide a better understanding of the methodology, we begin with a brief review of the general process of MAE. We then present our supervised attention-driven masking strategy, SAM, and discuss its application to the fine-tuning phase of MSMAE. 
Fig. \ref{fig4} illustrates the specific pipeline of MSMAE during the pre-training and fine-tuning phases, where SAM plays a crucial role in both phases.

\subsection{Preliminaries}

MAEs \cite{he2022masked} provide an efficient training scheme for vision transformers. While they are self-supervised, MAEs perform as well as or even outperforms supervised algorithms in several vision tasks. MAEs work mainly by using a high-ratio random masking strategy and an asymmetric encoder-decoder architecture.

Given an input image ${X} \in {{\mathbb{R}}^{{H} \times {W} \times 3}}$, the vision transformer first embeds it into $N + 1$ patches (include class patch) of size $p$ and adds positional embedding to it, where ${N} = {HW}/{{p}^2}$. Then MAE applies a random masking of as much as 75\% to all patches. 
Finally, the unmasked patches are encoded by a vision transformer and fed into a lightweight decoder along with learnable masks to recover the masked patches. MAE usually adopts the $L_2$ loss to recover the pixel information of the masked patches:
\begin{equation}
    {{\mathcal{L}}_{{con}}} = ||{G} - {Y}|{|_2},
  \label{eq1}
\end{equation}
where $G$ and $Y$ denote the RGB pixels of the masked patches and the predicted values of the learnable masks, respectively.

\subsection{Supervised Attention-Driven Masking Strategy}

Fig. \ref{fig1} shows that the use of a random masking strategy is likely to result in diseased tissues being missed by the model, as they can be subtle in medical images. 
Therefore, in medical image classification, it is necessary for the masked autoencoder to precisely pinpoint the diseased tissue so that the medical image can be accurately masked.
To address this issue, we propose supervised attention-driven masking (SAM). Specifically, SAM is divided into two core parts: one is responsible for supervising classification task, and the other is responsible for performing masking. 
The supervised classification task help the network to pinpoint the lesion areas in the medical images thus facilitating the masking strategy to achieve more accurate masking.

\begin{figure}[h]
  \centering
  \vspace{-2mm}
  \setlength{\abovecaptionskip}{1mm} 
  \setlength{\belowcaptionskip}{-2mm}
  \includegraphics[width=0.85\linewidth]{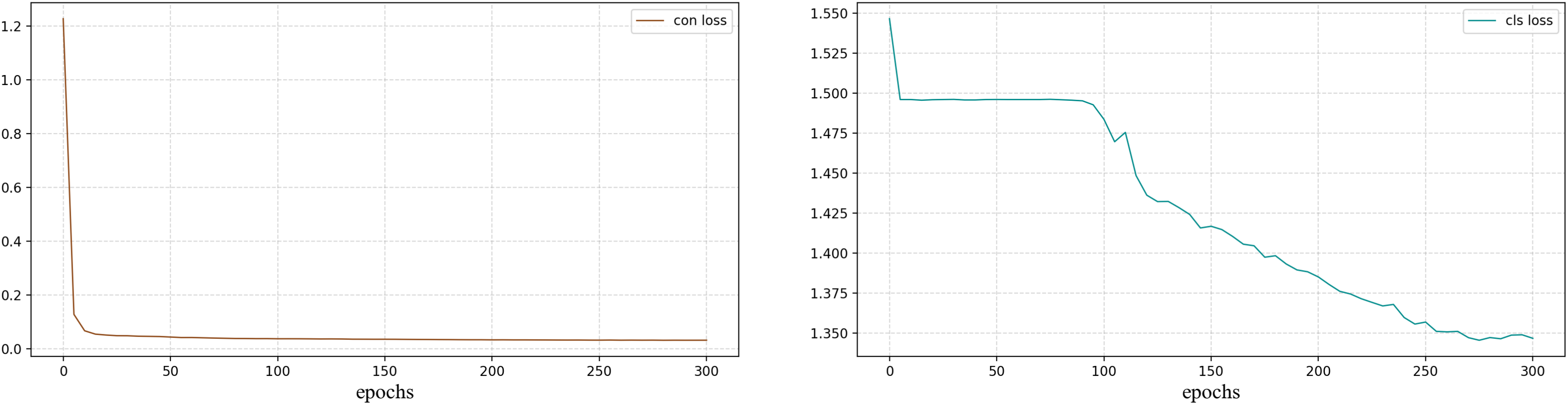}
   \caption{MSMAE image reconstruction loss (con loss) and classification loss (cls loss) in the pre-training phase on the HAM10000 dataset.}
   \label{fig5}
\end{figure}

\paragraph{Supervised Classification Task.} Apart from the image recovery task, the class token of MSMAE is required to perform the image classification task in the pre-training phase. 
Since the class token is not required to participate in the image recovery task during the pre-training phase, the supervised image classification task does not conflict with the image recovery task (see Fig. \ref{fig5}). 
Specifically, we perform a supervised classification task on the class token output by the encoder with the classification header, which is expressed as:
\begin{equation}
  \begin{aligned}
  &\left[{cls,e_{vis}}\right] = {\rm{E}}([cls, {x}_{vis}]),\\
  &{o} = {\rm{softmax}}({\rm{MLP}} \left(cls \right) ),\\
  &{{\mathcal{L}}_{{cls}}} =  - \frac{1}{{B}}\sum\limits_{{i} = 1}^{B} {{{t}_{i}}{\rm{log}}({{o}_{i}})}, 
\end{aligned}
\label{eq2}
\end{equation}
where $\rm{E}$ denotes the vision transformer encoder. ${{x}_{{vis}}}$, $cls$, and ${{e}_{{vis}}}$ represent visible tokens, class token, and unmasked patch encoding, respectively. $\left[ \cdot \right]$ indicates a merge operation.
${\rm{MLP}}$ denotes the classification head of the multilayer perceptron \cite{rumelhart1986learning}. ${\rm{softmax}}\left(\cdot\right)$ is the softmax activation function \cite{shannon1948mathematical}. $B$ and $t$ are the batch size and the ground truth, respectively.

The total loss in the pre-training process is:
\begin{equation}
  {{\mathcal{L}}_{{total}}} = {{\mathcal{L}}_{{con}}} + \lambda {{\mathcal{L}}_{{cls}}},
\label{eq3}
\end{equation}
where $\lambda $ is a hyperparameter. $\lambda $ is set to 0.1, if not specifically declared.

\paragraph{Masking Strategy.} Given the input ${z} \in {{\mathbb{R}}^{{(N} + {1)} \times {C}}}$, the affinity matrix calculation in the multi-head self-attention mechanism for each block in the vision transformer is denoted as:
\begin{equation}
  \begin{aligned}
      &{q} = {{W}_{q}}{z}, {k} = {{W}_{k}}{z},\\
      &{{a}_{i}} = {\rm{softmax}}({{q}_{i}}{{k}_{i}}/\sqrt {d} ), {i} \in [1,2,...,{h}],
\end{aligned}
\label{eq4}
\end{equation}
where ${{W}_{q}}{,}{{W}_{k}} \in {{\mathbb{R}}^{{C} \times {C}}}$ are two projection matrices. $d$ and $h$ refer to the embedding dimension and the number of attention heads, respectively.

After a period of pre-training, we use the vision transformer to encode the medical images and obtain the affinity matrix for the last layer block:
\begin{equation}
  \begin{aligned}
    {a} = \frac{1}{{h}}\sum\limits_{{i} = 1}^{h} {{{a}_{i}}} ,
\end{aligned}
\label{eq5}
\end{equation}
Then, we proceed to extract the constituent elements ${{a}_{{cls}}}$ (we also refer to it as the masking weight) of the rows that correspond to the class token, 
which have undergone supervised training, from the affinity matrix and subsequently subject them to a normalization \cite{ioffe2015batch} process (note that we have removed the element corresponding to the class token from the ${{a}_{{cls}}}$). 
Subsequently, the shape of ${{a}_{{cls}}}$ is changed to $({h}/{p}) \times ({w}/{p})$ and bilinearly interpolated back to the size of the original image.

\begin{figure}[t]
  \vspace{-2mm}
  \centering
  \setlength{\belowcaptionskip}{-4mm}
  \includegraphics[width=0.9\linewidth]{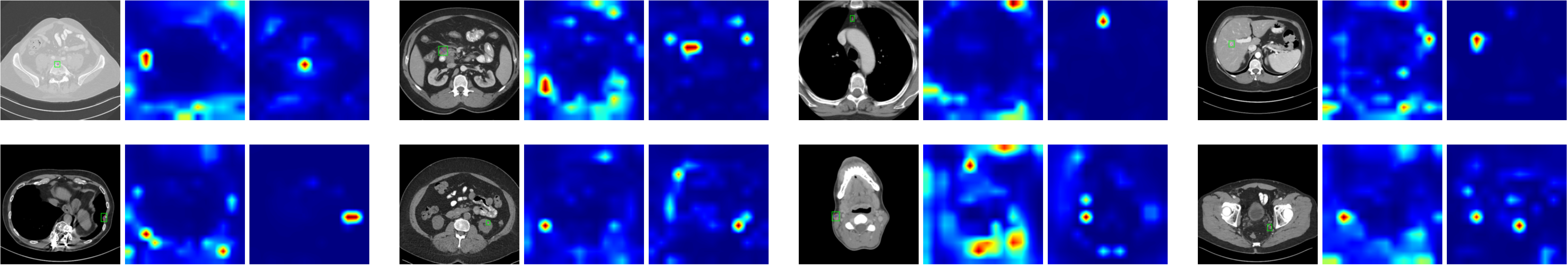}
   \caption{Masking weights of different algorithms in the pre-training phase.
   From left to right are the input image, the masking weights of the AMT and our SAM masking weights.
   The image inside the \textcolor{green}{green box} indicates the lesion-related human tissue. Please zoom in to see the details better.}
   \label{fig6}
\end{figure}

We visualize and compare SAM with other attention-based masking strategies in Fig. \ref{fig6}. Obviously, the ${{a}_{{cls}}}$ obtained by supervised training pinpoints the human tissue in the lesioned region.

Finally, we perform accurate masking and throwing of medical images using the ${{a}_{{cls}}}$ obtained from supervised training in the manner of AMT \cite{gui2022good}. Specifically, we first normalize and crop the ${{a}_{{cls}}}$ so that its elements correspond to the input image pixels individually. 
This masking weight is then transformed into patches and the index of the masking weight is sampled $N$ times:
\begin{equation}
  \begin{aligned}
    {J} = {\rm{sample}}({{a}_{cls}}),
\end{aligned}
\label{eq6}
\end{equation}
where the ${\rm{sample}}\left(\cdot\right)$ is a sampling function that tends to rank higher weights at the top and lower weights at the bottom and return their corresponding indexes. ${J} \in {{\mathbb{R}}^{N}}$ indicates the index after sampling.

The image tokens $x$ are then masked and thrown according to $J$:
\begin{equation}
  \begin{aligned}
      &{mask} = {J}[:{N} \times {r}],\\
      &{throw} = {J}[{N} \times {r:N} \times \left( {r} + t \right)],\\
      &{vis} = {J}[{N} \times \left({r} + t \right):],\\
      &{{x}_{{vis}}},{{x}_{{mask}}} = {x}[{vis}],{x}[{mask}],
\end{aligned}
\label{eq7}
\end{equation}
where $r$ and $t$ represent the masking ratio and throwing ratio respectively. $mask$, $throw$ and $vis$ represent the masked tokens index, the throw tokens index, and the visible tokens index respectively. $x_{mask}$ represent masked tokens.

\subsection{SAM Fine-tuning Schedule}

\begin{figure}[h]
  \centering
  \vspace{-3mm}
  \setlength{\abovecaptionskip}{1mm} 
  \setlength{\belowcaptionskip}{-2mm}
  \includegraphics[width=1\linewidth]{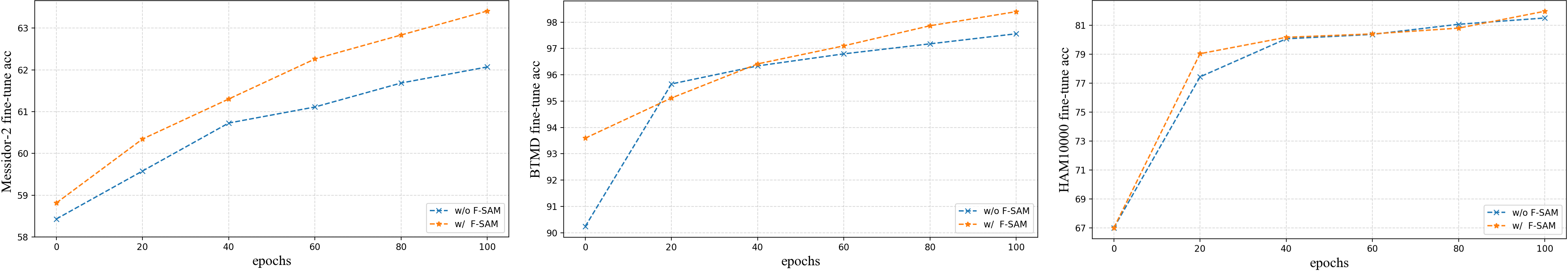}
   \caption{Differences in the accuracy of fine-tuning phase MSMAE with and without SAM on Messidor-2, BTMD, and HAM10000 datasets.}
   \label{fig7}
\end{figure}
Inconsistencies in MAE's pre-training and fine-tuning due to differences in inputs can cause performance and efficiency degradation in medical image classification task. To solve this problem, we extend SAM to the fine-tuning phase of MSMAE.

\paragraph{Fine-tuning Schedule.} During the fine-tuning phase, MSMAE continues to mask the medical images according to SAM from the pre-training phase. 
SAM makes the encoder input of MSMAE consistent between the fine-tuning phase and the pre-training phase thus improving the performance and efficiency of medical image classification task. 
This is specifically expressed as:
\begin{equation}
  \begin{aligned}
      &{{x}_{{vis}}},{{x}_{{mask}}} = {\rm{SAM}}({x, }{{a}_{{cls}}}),\\
      &[{cls,e_{vis}}] = {\rm{E}}([cls, {x}_{vis}]),\\
      &{o} = {\rm{softmax}}({\rm{MLP}}({cls})),\\
      &{{\mathcal{L}}_{{cls}}} =  - \frac{1}{{B}}\sum\limits_{{i} = 1}^{B} {{{t}_{i}}{\rm{log}}({{o}_{i}})} ,
\end{aligned}
\label{eq8}
\end{equation}
where ${\rm{SAM}}( \cdot )$ is the supervised attention-driven masking strategy. Fig. \ref{fig7} shows that the adoption of SAM for the fine-tuning schedule does not weaken the performance of MSMAE in classifying medical image.

For the testing phase, we employ all image tokens for testing.

\begin{figure}[h]
  \vspace{-2mm}
  \centering
  \setlength{\abovecaptionskip}{1mm} 
  \setlength{\belowcaptionskip}{-5mm}
  \includegraphics[width=1\linewidth]{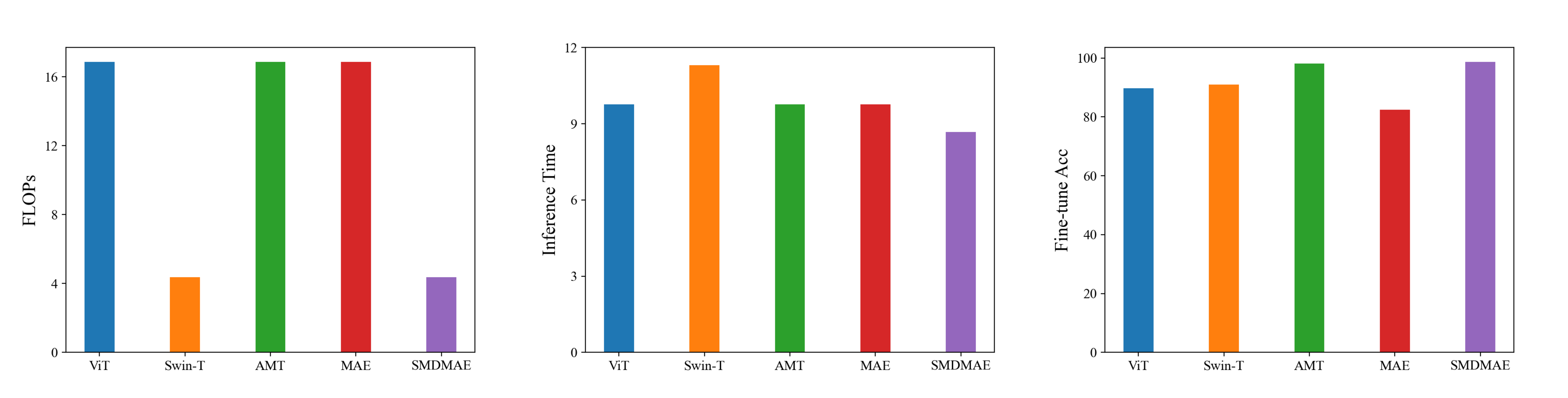}
   \caption{The results of FLOPs, inference time, and fine-tuning accuracy of different algorithms for the fine-tuning phase on the BTMD dataset.}
   \label{fig3}
\end{figure}

\paragraph{Discussion.} We consider that SAM is working in the fine-tuning phase for two main reasons. Firstly, the SAM in the fine-tuning phase successfully solved the problem of inconsistency between the MAE pre-training and fine-tuning phases. 
This prevents issues such as model performance degradation and over-fitting that may result from inconsistency. 
Secondly, similar to the dropout algorithm \cite{srivastava2014dropout}, SAM has a higher probability to mask the tokens associated with the lesion region, increasing the difficulty of fine-tuning and thus indirectly improving the diagnostic quality of the model utilizing all image tokens during testing.

\section{Experiments}
\subsection{Settings}

\paragraph{Datasets.} We validate the diagnostic capabilities of MSAME on several medical image classification tasks including diabetic retinopathy (DR), brain tumor (BT), and skin cancer. 
We perform the DR, BT, and skin cancer image classification tasks on the Messidor-2, BTMD, and HAM10000 datasets. We use the Breast Ultrasound Images (BUSI) Dataset for the medical image segmentation task. See Appendix \textcolor{red}{A} for more dataset settings.

\paragraph{Baseline Approaches.} In this section, we compare the MSMAE with supervised and self-supervised algorithms, respectively. We choose the vision transformer (ViT), swin transformer (Swin-T), and ResNet50 as the supervised algorithm baselines. 
For the self-supervised algorithms, we choose MAE, SimMIM, and AMT, which employ masking autoencoders, as well as MoCo v3, SimcLR, and contrast cropping algorithms from contrastive learning for comparison.

\paragraph{Implementation Details.} During the pre-training phase, we adopt AdamW \cite{loshchilov2017decoupled} with a base learning rate of 1e-3 as the optimizer for 300 epochs. The learning rate schedule is the cosine decay schedule. 
We use the batch size of 64. For SAM, we choose the masking rate of 0.45 and the throwing rate of 0.3. The warmup epochs default to 40 and the masking weight are updated every 40 epochs. Before the warmup epochs, we employ a random masking strategy for training. 
The rest of the pre-training settings are consistent with those of MAE. 
For the fine-tuning phase, we use the same SAM settings as in the pre-training phase and updated the masking weight every 20 epochs. Other settings are consistent with those of MAE fine-tuning. See Appendix \textcolor{red}{B} for specific settings. 

\subsection{Comparison with Supervised Algorithms}

  \begin{table}[h]
    \setlength{\abovecaptionskip}{2mm} 
    \setlength{\belowcaptionskip}{-1mm}
    \centering
    \caption{Comparison results of MSMAE with several SOTA algorithms on various medical image classification tasks.}
    \label{tab1}
    \setlength\tabcolsep{5pt}
    \tiny
    \renewcommand\arraystretch{1.1}
    \begin{tabular}{c|c|c|c|c|c}
    \hline\thickhline
    \rowcolor{mygray}
     &   & &  \multicolumn{3}{c}{ \textbf{Dataset} }  \\
     \cline{4-6}
     \rowcolor{mygray}
      \multirow{-2}*{Type}&\multirow{-2}*{Method}  &\multirow{-2}*{Backbone} & Messidor-2 & BTMD & HAM10000  \\
      \hline\hline

      Contrastive     & MoCo v3 \cite{chen2021empirical}                      & ViT-B \cite{dosovitskiy2020image}       & 61.30              & 89.01             & 76.24 \\
      self-supervised & Contrastive Crop (MoCo v2) \cite{peng2022crafting}              & ResNet50 \cite{he2016deep}  & 59.77              & 95.96             & 75.31  \\
      learning        & SimcLR \cite{chen2020simple}                      & ResNet50 \cite{he2016deep}   & 59.75              & 90.62             & 75.00 \\

      \hline
    MIM               & MAE   \cite{he2022masked}              & ViT-B \cite{dosovitskiy2020image}      & 60.54              & 82.46             & 75.01  \\
    self-supervised   & SimMIM   \cite{xie2022simmim}        & Swin-T \cite{liu2021swin}     & 58.43              & 70.33             & 70.08 \\
    learning          & AMT (MAE) \cite{gui2022good}         & ViT-B \cite{dosovitskiy2020image}      & 60.92              & 98.17             & 81.53 \\
    \hline
                      & VIT-B  \cite{dosovitskiy2020image}           & ViT-B \cite{dosovitskiy2020image}      & 59.77             & 89.78             & 79.54 \\
    Supervised        & ResNet50 \cite{he2016deep}      & ResNet50 \cite{he2016deep}   & 58.43             & 69.18             & 67.01 \\
    learning          & Swin-T \cite{liu2021swin}        & Swin-T \cite{liu2021swin}     & 58.43             & 91.00                & 81.70 \\ 
    \cline{2-6}          
                      & \textbf{Ours}                 & ViT-B \cite{dosovitskiy2020image}      & {\bbetter{25}{15}{\textbf{63.41}}{2.11}}   & {\bbetter{25}{15}{\textbf{98.39}}{0.22}}  & {\bbetter{25}{15}{\textbf{81.97}}{0.27}} \\
    
    \hline\thickhline
    \end{tabular}
    \vspace{-2mm}
    \end{table}

\paragraph{Results.} Tab. \ref{tab1} summarizes the results of the evaluation of MSMAE and a supervised algorithm on three medical image classification tasks. 
The results show that MSMAE outperforms the supervised algorithm baseline in all three tasks, highlighting the effectiveness of SAM-based image recovery in the pre-training phase, allowing MSMAE to accurately capture the human tissue representation of lesion-related regions. 
Moreover, SAM enhances the difficulty of fine-tuning of MSMAE and improves the diagnostic quality of the model.
These results demonstrate the significant potential of MSMAE for medical image analysis tasks, particularly in cases where large amounts of labeled data are not readily available.

\subsection{Comparison with Self-supervised Algorithms}

\begin{figure}[h]
  \centering
  \setlength{\abovecaptionskip}{2mm} 
  \includegraphics[width=1\linewidth]{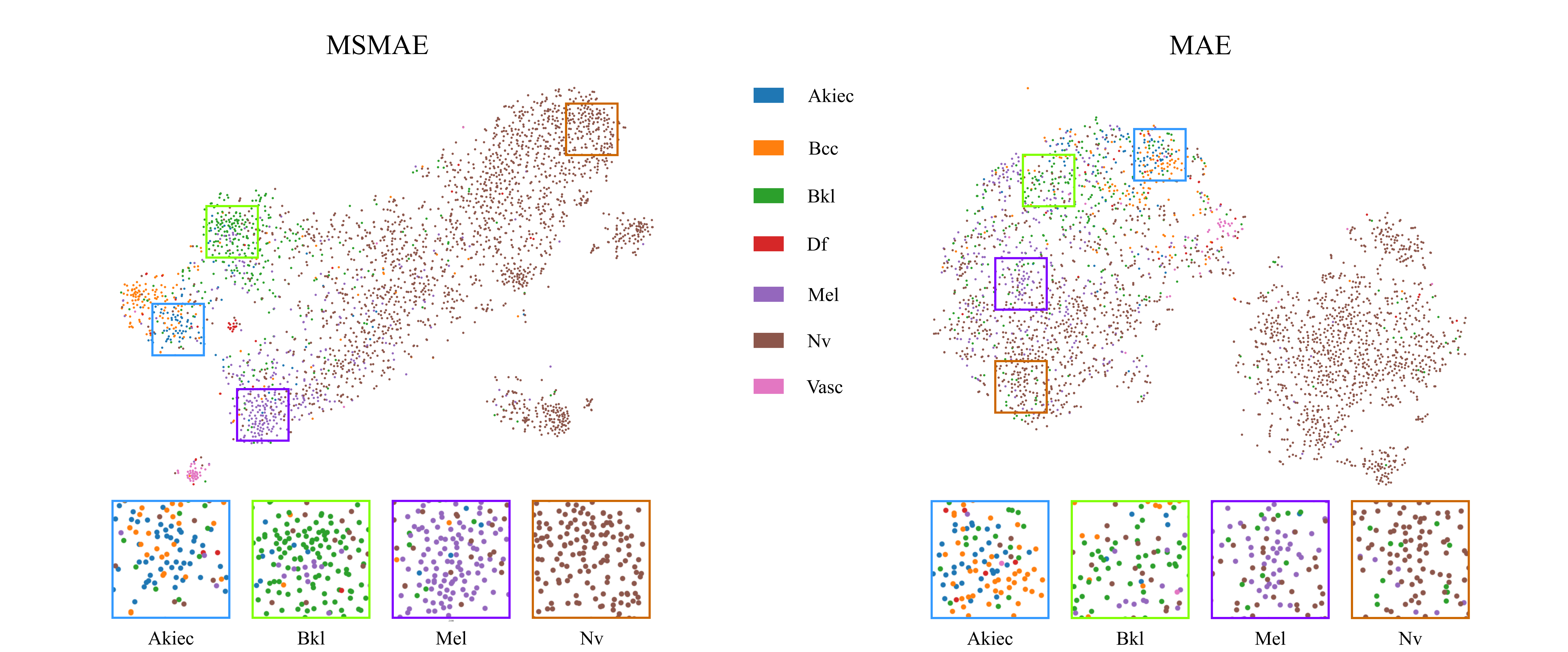}
   \caption{Visualization of MSMAE and MAE representations on HAM10000. We visualize the representation of MSMAE and MAE using tSNE \cite{van2008visualizing}. Compared to MAE, MSMAE achieves better category separation at HAM. 
   Meanwhile, in the magnified image below, we find that MSMAE achieves a better single-class aggregation. This demonstrates that MSMAE achieves better representation learning on HAM10000. 
   Actinic keratoses and intraepithelial carcinoma (Akiec), basal cell carcinoma (Bcc), benign keratosis-like lesions (Bkl), dermatofibroma (Df), melanoma (Mel), melanocytic nevi (Nv) and vascular lesions (Vasc).}
   \label{fig8}
   \vspace{-1mm}
\end{figure}

\paragraph{Results.} The effectiveness of the supervised learning and SAM-based masking strategies employed in MSMAE are evaluated in the pre-training phase by comparison with other self-supervised algorithms.
Tab. \ref{tab1} presents experimental results that show MSMAE outperforms SOTA self-supervised algorithms in several medical image classification tasks. These results highlight the importance of precise masking strategies in medical image classification task. 
Moreover, the strong generalization ability of MSMAE is demonstrated: it is not limited to a single type of medical image classification task, which is beneficial in addressing the scarcity of medical diagnostic equipment and professionals. 
Furthermore, the representation learned by MSMAE and other algorithms are visualized in Fig. \ref{fig8}, revealing that MSMAE produces higher-quality feature learning and class separation on medical image classification datasets. 
These results suggest that the proposed MSMAE with the SAM-based masking strategy is an effective approach for medical image classification task.

\begin{figure}[h]
  \centering
  \setlength{\abovecaptionskip}{1mm} 
  \setlength{\belowcaptionskip}{-3mm}
  \includegraphics[width=0.9\linewidth]{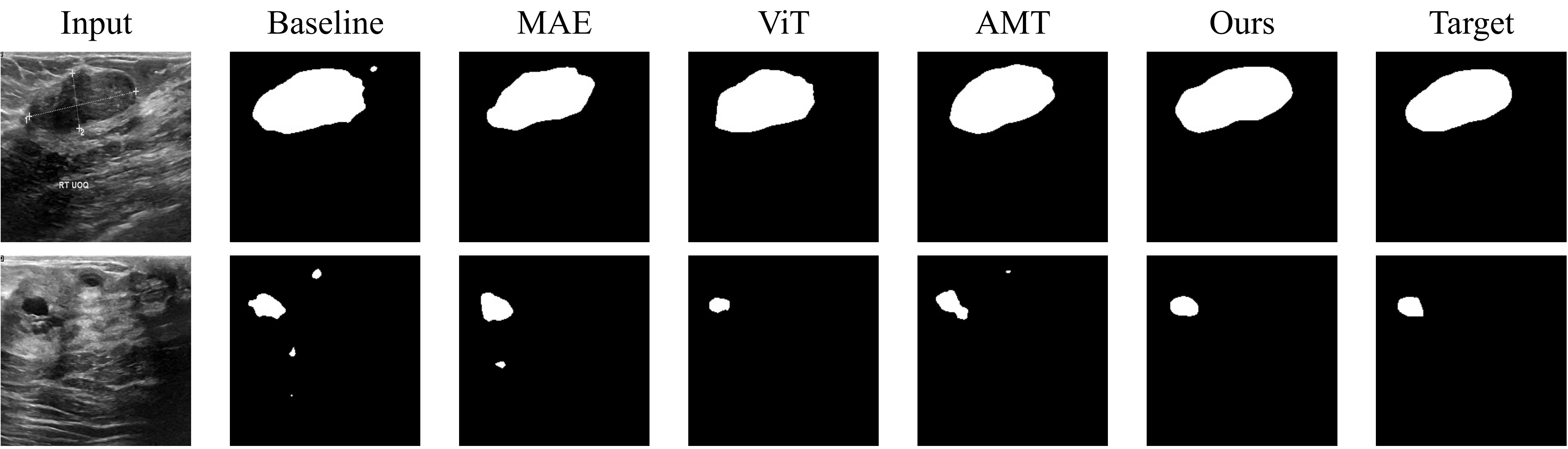}
   \caption{Visualization of semantic segmentation results for breast cancer.}
   \label{fig9}
\end{figure}
\subsection{Medical Semantic Segmentation}

  To further validate the transferability of MSMAE, we pre-trained MSMAE on the BUSI dataset and verify its performance on the breast cancer segmentation task. 
  The specific experimental setup and the breast cancer classification experiments associated with BUSI are shown in Appendix \textcolor{red}{C}.

After pre-training, ViT-B \cite{dosovitskiy2020image} from MSMAE and other algorithms were used as encoders for SETR \cite{zheng2021rethinking} to fine-tune and test the BUSI dataset. 
We report the performance of MSMAE and other algorithms on the semantic segmentation task for breast cancer in Tab. \ref{tab2} and visualize it in Fig. \ref{fig9}.

\begin{table}[h]
  \setlength{\abovecaptionskip}{1mm} 
  \setlength{\belowcaptionskip}{-2mm}
  \tiny
  \centering
  \caption{Semantic segmentation results on BUSI. SETR serves as the semantic partitioning framework for all algorithms in the table. Baseline indicates that SETR does not use the ViT initialized by the pre-trained model.}
  \label{tab2}
  \renewcommand{\arraystretch}{1.1}
  \setlength{\tabcolsep}{1mm}{  
  \begin{tabular}{c|c|c|c|c}
      \hline\thickhline
      \rowcolor{mygray}
      \textbf{Method} & \textbf{Backbone}        & \textbf{Pre-training Epochs} & \textbf{F1 Score}                & \textbf{IoU} \\
      \hline\thickhline
    Baseline         &SETR \cite{zheng2021rethinking} & -                 &  56.15 & 43.33 \\
    \hline
    MAE \cite{he2022masked}    &SETR \cite{zheng2021rethinking}    & 300               &   57.09 & 43.95 \\
    AMT (MAE) \cite{gui2022good}  &SETR \cite{zheng2021rethinking}   & 300               &   56.17 & 43.07 \\
    VIT-B \cite{dosovitskiy2020image}  &SETR \cite{zheng2021rethinking}   & 300               &   56.22	& 42.66 \\
    \hline
    \textbf{Ours}     &SETR \cite{zheng2021rethinking}        & 300               & {\bbetter{25}{15}{\textbf{60.66}}{3.57}}             & {\bbetter{25}{15}{\textbf{45.53}}{1.58}} \\
    \hline\thickhline
    \end{tabular}}%
    \vspace{-2mm}
  \end{table}

It is noticed that with the same ViT backbone, our method achieves better a F1 score and Intersection over Union (IoU). 
Moreover, our breast cancer semantic segmentation results are closer to the ground truth than other algorithms. 
This shows that MSMAE has great potential for medical semantic segmentation and demonstrates our approach's stronger transferability for medical processing.

\subsection{Ablations}

We utilized Messidor-2 as the benchmark for these experiments and MAE as the baseline model. See Appendix \textcolor{red}{C} for more ablation experiments.

\paragraph{SAM.} Tab. \ref{tab3} demonstrates the influence of SAM on MSMAE's performance in medical image classification task. 
We explored the effect of three masking strategies on the pre-training phase of the MAE medical image classification task. 
The best diagnostic result was achieved at the baseline with SAM pre-training. This indicates that our SAM is more suitable for processing medical images than other masking strategies.

\begin{table}[h]
  \vspace{-4mm}
  \setlength{\abovecaptionskip}{1mm} 
  \tiny
  \centering
  \caption{Results of ablation experiments of masking strategies and fine-tuning schedule. P-RM, P-AMT and P-SAM represent the random masking strategy, 
  attention-driven masking strategy and supervised attention-driven masking strategy in the pre-training phase, respectively. 
  F-SAM indicates a supervised attention-driven masking strategy for the fine-tuning phase.}
  \label{tab3}
  \renewcommand{\arraystretch}{1.1}
  \setlength{\tabcolsep}{1mm}{  
  \begin{tabular}{c|cccc|c|c|c}
      \hline\thickhline
      
      \rowcolor{mygray}
       &            \multicolumn{4}{c|}{\textbf{Module}}                  &       \multicolumn{3}{c}{\textbf{Fine-tuning}}              \\
       \cline{2-8}
      \rowcolor{mygray}
      \multirow{-2}{*}{\textbf{Setting}} & P-RM & P-AMT & P-SAM & F-SAM  & Top-1 Acc (\%)   & FLOPs             & Inference time \\
      \hline\hline
    Baseline          & \checked              &                   &                   &                   & 60.54             & 16.86G            & 9.765ms \\
    \uppercase\expandafter{\romannumeral2}                 &                   & \checked              &                   &                   & 60.92             & 16.86G            & 9.765ms \\
    \uppercase\expandafter{\romannumeral3}                 &                   &                   & \checked              &                   & 62.07             & 16.86G            & 9.765ms \\
    \uppercase\expandafter{\romannumeral4}                 & \checked              &                   &                   & \checked              & 60.58             & 4.37G             & 8.671ms \\
    \uppercase\expandafter{\romannumeral5}                 &                   & \checked              &                   & \checked              & 61.39             & 4.37G             & 8.671ms \\
    \uppercase\expandafter{\romannumeral6}                 &                   &                   & \checked              & \checked              & \underline{63.41}             & \underline{4.37G}             & \underline{8.671ms} \\
    \hline\thickhline
    \end{tabular}}%
    \vspace{-2mm}
  \end{table}

\paragraph{Fine-tuning Schedule}.  We performed an ablation study of the fine-tuning schedule to observe whether our SAM-based fine-tuning schedule worked. The experimental results are shown in Tab. \ref{tab3}. 
Both AMT and SAM pre-trained baselines with SAM fine-tuning were employed to achieve significant improvements. 
We think the reason for the insignificant improvement of the baseline paired with SAM fine-tuning by pre-training with a random masking strategy may be the difference between random masking and attention-based masking strategies. 
In addition, SAM's fine-tuning schedule drastically reduces the computation and inference time of the baseline in the fine-tuning phase, thus improving training efficiency.

\section{Conclusion}

In this paper, we propose a novel masked autoencoder MSMAE designed for medical image classification tasks. 
MSMAE can perform accurate masking of medical images driven by supervised attention during the pre-training phase. Thus, it can effectively learn lesion-related human representations. 
It is worth noting that our method also extends this masking strategy to the fine-tuning phase. 
This prevents the issue of inconsistency between the pre-training and fine-tuning phases of MSMAE and enhances the fine-tuning difficulty of MSMAE. 
As a result, MSMAE improves the quality of medical image classification task with great efficiency. 
We have demonstrated the excellent medical diagnostic capabilities of MSMAE through extensive experiments on various medical image classification tasks. 
This suggests that MSMAE is a promising solution to the shortage of medical professionals and equipment.

\clearpage

\appendix

\section*{Appendix}

\section{Datasets}

We used four publicly available medical datasets for our experiments: Messidor-2, Brain Tumor MRI Dataset, Breast Ultrasound Images Dataset, and HAM10000.

\paragraph{Brain Tumor MRI.} Brain Tumor MRI Dataset (BTMD) \cite{amin2021brain} consists of 7023 magnetic resonance imaging (MRI) images of the human brain which have been categorized into four different classes: glioma, meningioma, pituitary, and no tumor. 
The dataset is composed of three datasets: The Br35H \cite{Ahmedbr35h} dataset provides normal images without tumors, glioma category images come from data on the figshare website, and other images from the Brain Tumor Classification (MRI) \cite{sartaj} dataset.
We randomly divided this dataset into a training set of 5712 images and a validation set of 1311 images.

\paragraph{Messidor-2.} Messidor-2 \cite{abramoff2016improved} is a public medical image dataset commonly used for the detection and diagnosis of diabetic retinopathy (DR). 
The dataset contains 1748 macula-centric fundus images annotated by specialist researchers and can be divided into five categories: No diabetic retinopathy, Mild non-proliferative retinopathy, Moderate non-proliferative retinopathy, Severe non-proliferative retinopathy, and Proliferative diabetic retinopathy. 
We divided 1744 of these images with labels into a training set and a validation set. The training set contains 1222 images and the validation set contains 522 images.

\paragraph{HAM10000.} HAM10000 \cite{codella2019skin} contains 10015 dermatoscopy images from different populations, including 7014 images in the training set and 3001 images in the validation set. 
Skin cancer cases contain typical categories in the field of pigmented lesions. It contains 7 different types of skin cancer as follows: Intra-Epithelial Carcinoma(AKIEC), Basal Cell Carcinoma(BCC), Benign Keratosis(BKL), Dermatofibroma(DF), Melanoma(MEL), Melanocytic Nevi(NV), and Vascular lesions(VASC) . 
The dermatological cases include the typical categories in the field of pigmented lesions, most of which have been tested by pathology, with a small number of cases verified by expert peer review or follow-up.

\paragraph{Breast Ultrasound Images Dataset.} Breast Ultrasound Images Dataset (BUSI) \cite{al2020dataset} include ultrasound scanned breast images of 600 women between the ages of 25 and 75. 
The dataset consists of 780 breast ultrasound images with an average size of 500$\times$500 pixels and each image has a corresponding tumor image (mask). 
The images were classified as normal, benign, and malignant. We randomly divide 547 images as the training set and 233 images as the validation set and perform the semantic segmentation task using the corresponding tumor images.

\section{Implementation Details}

\begin{table}[h]
  \centering
  \footnotesize
  \caption{MSMAE pre-training settings.}
  \label{tabs1}
  \setlength\tabcolsep{5pt}
  \renewcommand\arraystretch{1.1}
  \begin{tabular}{c|c}
  \hline\thickhline
  config	&value                  \\
  \hline
  optimizer	&AdamW \cite{loshchilov2017decoupled}    \\
  base learning rate &	1e-3   \\
  weight decay	&0.05    \\
  model	&   msmae\_vit\_base\_patch16    \\
  input size	&   224    \\
  batch size	&   64    \\
  learning rate schedule&cosine decay \cite{loshchilov2016sgdr}    \\
  training epochs	&300    \\
  warmup epochs	&40    \\
  mask ratio	&0.45    \\
  throw ratio	&0.30    \\
  ${{a}_{{cls}}}$ update interval	&40    \\
  augmentation	&random center cropping, random horizontal flipping    \\
  \hline\thickhline
  \end{tabular}
  \end{table}

  \begin{table}[h]
    \centering
    \footnotesize
    \caption{MSMAE fine-tuning settings.}
    \label{tabs2}
    \setlength\tabcolsep{5pt}
    \renewcommand\arraystretch{1.1}
    \begin{tabular}{c|c}
    \hline\thickhline
    config	&value                  \\
    \hline
    optimizer	&AdamW \cite{loshchilov2017decoupled}    \\
    base learning rate &	1e-3    \\
    weight decay	&0.05    \\
    layer-wise lr decay \cite{you2017large}	&0.75\\
    model	&   vit\_base\_patch16    \\
    input size	&   224    \\
    batch size	&   256    \\
    learning rate schedule&cosine decay \cite{loshchilov2016sgdr}    \\
    training epochs	&100    \\
    warmup epochs	&5    \\
    mask ratio	&0.45    \\
    throw ratio	&0.30    \\
    ${{a}_{{cls}}}$ update interval	&20    \\
    augmentation	&random center cropping, random horizontal flipping    \\
    \hline\thickhline
    \end{tabular}
    \end{table}

For all medical image classification experiments of MSMAE, we use ViT-Base \cite{dosovitskiy2020image} as the encoder. 
For the decoder of MSMAE, it consists of 8 vanilla transformer blocks with embedding dimension of 512 and number of attention heads of 16. 
In the pre-training phase, MSMAE data augmentation strategy employs a random center cropping with a scale of (0.2, 1.0) an interpolation of 3, and a random horizontal flipping.
In the fine-tuning phase, MSMAE uses the same data augmentation strategy as in the pre-training phase. 
All experiments were done on a single NVIDIA RTX 3090. And we use the PyTorch framework for all code. Implementation details can be found in Tab. \ref{tabs1} and Tab. \ref{tabs2}.

\section{More Experiments}

\subsection{Medical Semantic Segmentation Settings}

We transferred the pre-trained ViT to SETR for the breast cancer segmentation task. 
For SETR training we use binary cross entropy \cite{yi2004automated} and dice loss \cite{milletari2016v}. Specific semantic segmentation losses are as follows:
\begin{equation}
  {{\mathcal{L}}_{{seg}}} = {\rm{dice}}({o},{t}) + \lambda {{\rm{bce}}(o,t),}
\label{eqs1}
\end{equation}
where ${o}$ denotes the predicted result and ${t}$ denotes the ground truth. ${\rm{dice}}( \cdot )$ and ${\rm{bce}}( \cdot )$ denote the dice loss and binary cross entropy loss, respectively. 
For $\lambda $, we default to 0.5.

\begin{table}[t]
  \centering
  \footnotesize
  \caption{Breast cancer semantic segmentation settings.}
  \label{tabs3}
  \setlength\tabcolsep{5pt}
  \renewcommand\arraystretch{1.1}
  \begin{tabular}{c|c}
  \hline\thickhline
  config	&value                  \\
  \hline
  optimizer	&Adam \cite{kingma2014adam}    \\
  base learning rate &	1e-4    \\
  weight decay	&   1e-4    \\
  model	&   setr    \\
  input size	&   224    \\
  batch size	&   8    \\
  learning rate schedule&cosine decay \cite{loshchilov2016sgdr}    \\
  training epochs	&200    \\
  augmentation&	random rotation \\
  \hline\thickhline
  \end{tabular}
  \end{table}

\paragraph{Settings.} For the pre-training phase and breast cancer classification experiments we followed the previous experimental settings. 
For medical semantic segmentation experiments, we used the Adam optimizer with a learning rate of 1e-4 and a momentum of 0.9 for 200 epochs of fine-tuning. 
Batch size we set to 8. The learning rate schedule uses a cosine decay schedule. 
Besides, for the data augmentation strategy we choose random rotation. More experimental settings are shown in Tab. \ref{tabs3}.

\subsection{Breast Cancer Classification Experiment}

\begin{table}[h]
  \centering
  \footnotesize
  \caption{Results of medical image classification for breast cancer.}
  \label{tabs4}
  \setlength\tabcolsep{5mm}
  \renewcommand\arraystretch{1.1}
  \begin{tabular}{c|c|c|c}
  \hline\thickhline
  \rowcolor{mygray}
   & & & \\
  \rowcolor{mygray}
  \multirow{-2}*{Type}              & \multirow{-2}*{Method}            & \multirow{-2}*{Backbone}          & \multirow{-2}*{BUSI} \\
  \hline\hline
 Self-supervised  & MAE \cite{he2022masked}              & ViT-B \cite{dosovitskiy2020image}         & 72.96 \\
 learning        & AMT (MAE) \cite{gui2022good}        & ViT-B \cite{dosovitskiy2020image}      & 66.09 \\
                    \hline
  Supervised & VIT-B \cite{dosovitskiy2020image}      & ViT-B  \cite{dosovitskiy2020image}     & 74.25 \\
  \cline{2-4}
  learning        & \textbf{Ours}              & ViT-B  \cite{dosovitskiy2020image}                        & {\bbetter{25}{15}{\textbf{77.68}}{3.43}} \\
  \hline\thickhline
  \end{tabular}
  \end{table}

\paragraph{Results.} Tab. \ref{tabs4} shows the performance of our method on the breast cancer classification task. 
It is worth noting that our MSMAE achieves an Top-1 accuracy of 77.68 on BUSI. This surpasses the MAE and supervised baseline of +4.72 and +3.43, respectively. 

\subsection{More Breast Cancer Semantic Segmentation Results}

\begin{figure}[h]
  \centering
  \includegraphics[width=1\linewidth]{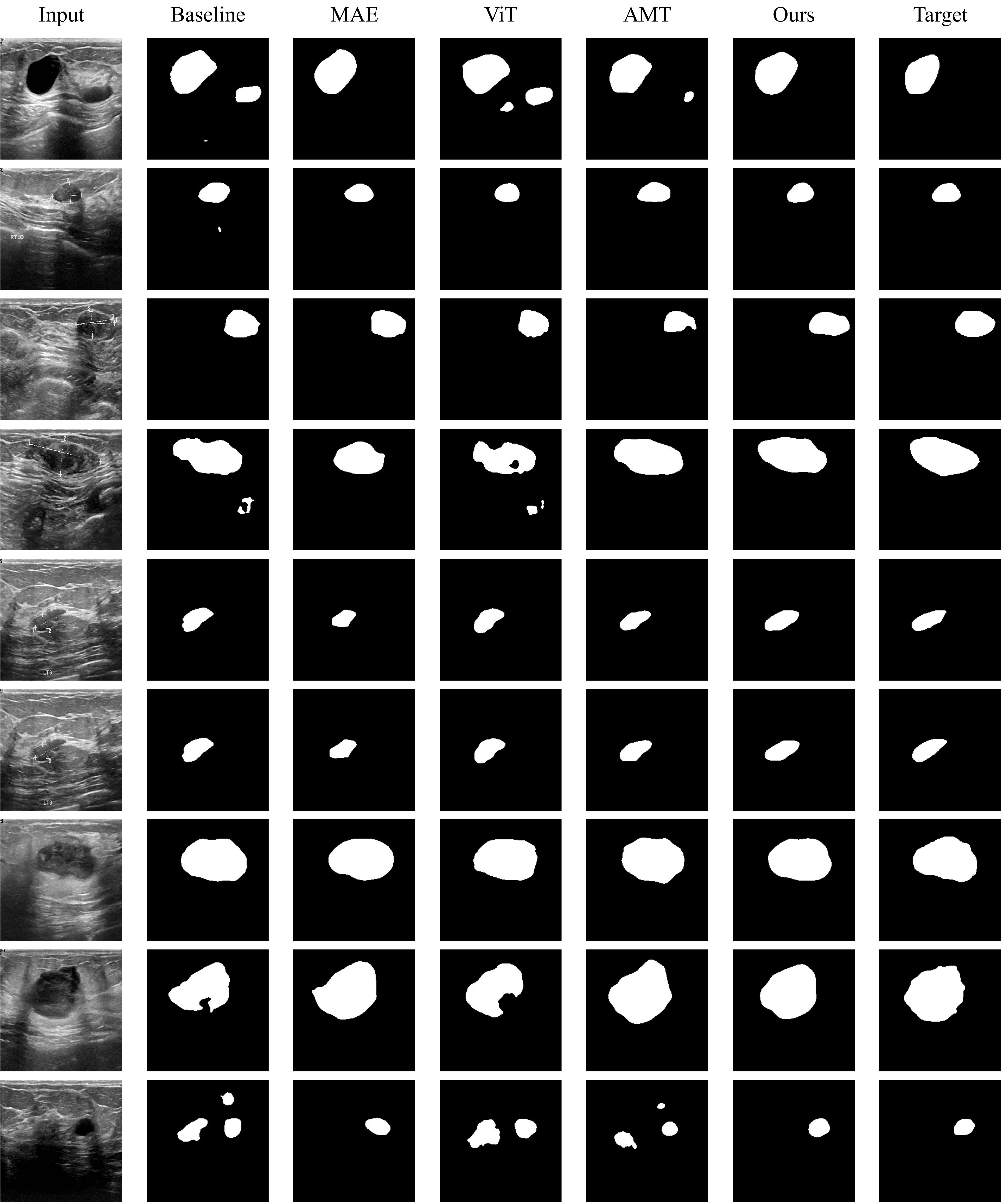}
   \caption{More results on semantic segmentation of breast cancer on BUSI.}
   \label{figs1}
\end{figure}

Moreover, we show more breast cancer semantic segmentation results in Fig. \ref{figs1}.

\subsection{Hyperparametric Ablation Study}

Tab. \ref{tabs5} presents the results of the ablation experiments conducted to investigate the effects of various hyperparameters on the performance of MSMAE in medical image classification tasks.

\begin{table}[h]
  \captionsetup{position=bottom}
  \vspace{-.2em}
  \centering
  \caption{MSMAE ablation study of various hyperparameters. We report the accuracy of MSMAE on the DR task and the BT task.}
  \label{tabs5} \vspace{-.5em}
  \subfloat[
    \scriptsize\textbf{Pre-training batch size (p-bs).} The smaller pre-training batch size can improve the medical image classification accuracy of MSMAE.
  \label{tab:decoder_depth}
  ]{
  \centering
  \begin{minipage}{0.25\linewidth}{\begin{center}
  \tablestyle{4pt}{1.05}
  \scriptsize
  \begin{tabular}{x{24}x{24}x{24}}
    p-bs & DR Acc. & BT Acc. \\
  \shline
  64 & 63.41 & 98.39 \\
  128 & 62.87 & 97.70 \\
  256 & 62.98 & 97.77 \\
  \multicolumn{3}{c}{~}\\
  \end{tabular}
  \end{center}}\end{minipage}
  }
  \hspace{2.5em}
  \subfloat[
    \scriptsize\textbf{Fine-tuning batch size (f-bs).} Influence of fine-tuning batch size on the effectiveness of MSMAE medical image classification.
  \label{tab:decoder_width}
  ]{
  \begin{minipage}{0.25\linewidth}{\begin{center}
  \tablestyle{4pt}{1.05}
  \scriptsize
  \begin{tabular}{x{24}x{24}x{24}}
    f-bs & DR Acc. & BT Acc. \\
  \shline
  64 & 63.98 & 97.52 \\
  128 & 63.17 & 98.03 \\
  256 & 63.41 & 98.39 \\
  \multicolumn{3}{c}{~}\\
  \end{tabular}
  \end{center}}\end{minipage}
  }
  \hspace{2.5em}
  \subfloat[
    \scriptsize\textbf{Mask ratio (mr) and throw ratio (tr).} Accuracy of MSMAE on medical image classification tasks with different masking ratios and throw ratios.
  \label{tab:mask_token}
  ]{
  \begin{minipage}{0.25\linewidth}{\begin{center}
  \tablestyle{4pt}{1.05}
  \scriptsize
  \begin{tabular}{x{14}x{14}x{24}x{24}}
    mr & tr & DR Acc. & BT Acc. \\
  \shline
  75\% & 0\% & 62.93 & 97.58 \\
  45\% & 40\% & 63.04 & 97.93 \\
  45\% & 30\% & 63.41 & 98.39 \\
  30\% & 45\% & 63.02 & 98.16 \\
  \end{tabular}
  \end{center}}\end{minipage}
  }
  \\
  \centering
  \vspace{.3em}
  \subfloat[
    \scriptsize\textbf{Pre-training masking weights update interval (p-mwui). }Differences of different masking weights update intervals in the pre-training phase on MSMAE medical image classification tasks.
  \label{tab:mae_target} 
  ]{
  \begin{minipage}{0.25\linewidth}{\begin{center}
  \tablestyle{4pt}{1.05}
  \scriptsize
  \begin{tabular}{x{24}x{24}x{24}}
    p-mwui & DR Acc. & BT Acc. \\
  \shline
  20 & 62.27 & 97.25 \\
  40 & 63.41 & 98.39 \\
  60 & 63.55 & 98.03 \\
  80 & 63.06 & 97.94 \\
  \end{tabular}
  \end{center}}\end{minipage}
  }
  \hspace{2.5em}
  \subfloat[
    \scriptsize\textbf{Fine-tuning masking weights update interval (f-mwui). }Differences of different masking weights update intervals in fine-tuning phase on MSMAE medical image classification tasks.
  \label{tab:aug}
  ]{
  \centering
  \begin{minipage}{0.25\linewidth}{\begin{center}
  \tablestyle{4pt}{1.05}
  \scriptsize
  \begin{tabular}{x{24}x{24}x{24}}
    f-mwui & DR Acc. & BT Acc. \\
  \shline
  10 & 62.88 & 97.16 \\
  20 & 63.41 & 98.39 \\
  30 & 63.36 & 98.27 \\
  40 & 63.30 & 97.29 \\
  \end{tabular}
  \end{center}}\end{minipage}
  }
  \hspace{2.5em}
  \subfloat[
    \scriptsize\textbf{Global pool (gp). }During the fine-tuning phase, our MSMAE works when performing global pooling operations on all tokens.
  \label{tab:mask_types}
  ]{
  \begin{minipage}{0.25\linewidth}{\begin{center}
  \tablestyle{4pt}{1.05}
  \scriptsize
  \begin{tabular}{x{24}x{24}x{24}}
    gp & DR Acc. & BT Acc.\\
  \shline
  True & 63.41 & 98.39 \\
  False & 63.03 & 98.47 \\
  \multicolumn{3}{c}{~}\\
  \multicolumn{3}{c}{~}\\
  \end{tabular}
  \end{center}}\end{minipage}
  }
  \vspace{-.1em}
  \end{table}

\section{Implementation of MSMAE Fine-tuning}

\begin{algorithm}[t]
  \caption{Pseudo-Code of MSMAE fine-tuning.}
  \label{algs1}
  \definecolor{codeblue}{rgb}{0.25,0.5,0.5}
  \definecolor{codekw}{rgb}{0.85, 0.18, 0.50}
  \begin{lstlisting}[language=python]
   # obtain mask_weights 
   mask_weights = update_mask(model, data_loader_eval)
   # perform SAM masking strategy fine-tuning on the input image.
   # masking and throwing according to our masking weights in the way of amt.
   x = amt_masking_throwing(x, mask_ratio, throw_ratio, mask_weights)
   cls_token = cls_token + pos_embed[:, :1, :]
   cls_tokens = cls_token.expand(x.shape[0], -1, -1)
   x = torch.cat((cls_tokens, x), dim=1)
   x = pos_drop(x)
   for i, blk in enumerate(blocks):
       x = blk(x)
   # global pool without cls token
   x = x[:, 1:, :].mean(dim=1)
   outcome = fc_norm(x)
   outcome = head(outcome)
   return outcome
   
    
  \end{lstlisting}
  \end{algorithm}

Algorithm \ref{algs1} shows the specific implementation details of the MSMAE fine-tuning. Our MSMAE updates the masking weights every 20 epochs. 
Thanks to the introduction of the SAM strategy, our MSMAE is able to be so efficient in the fine-tuning phase.

\section{More Visualization Results}

\begin{figure}[h]
  \centering
  \includegraphics[width=1\linewidth]{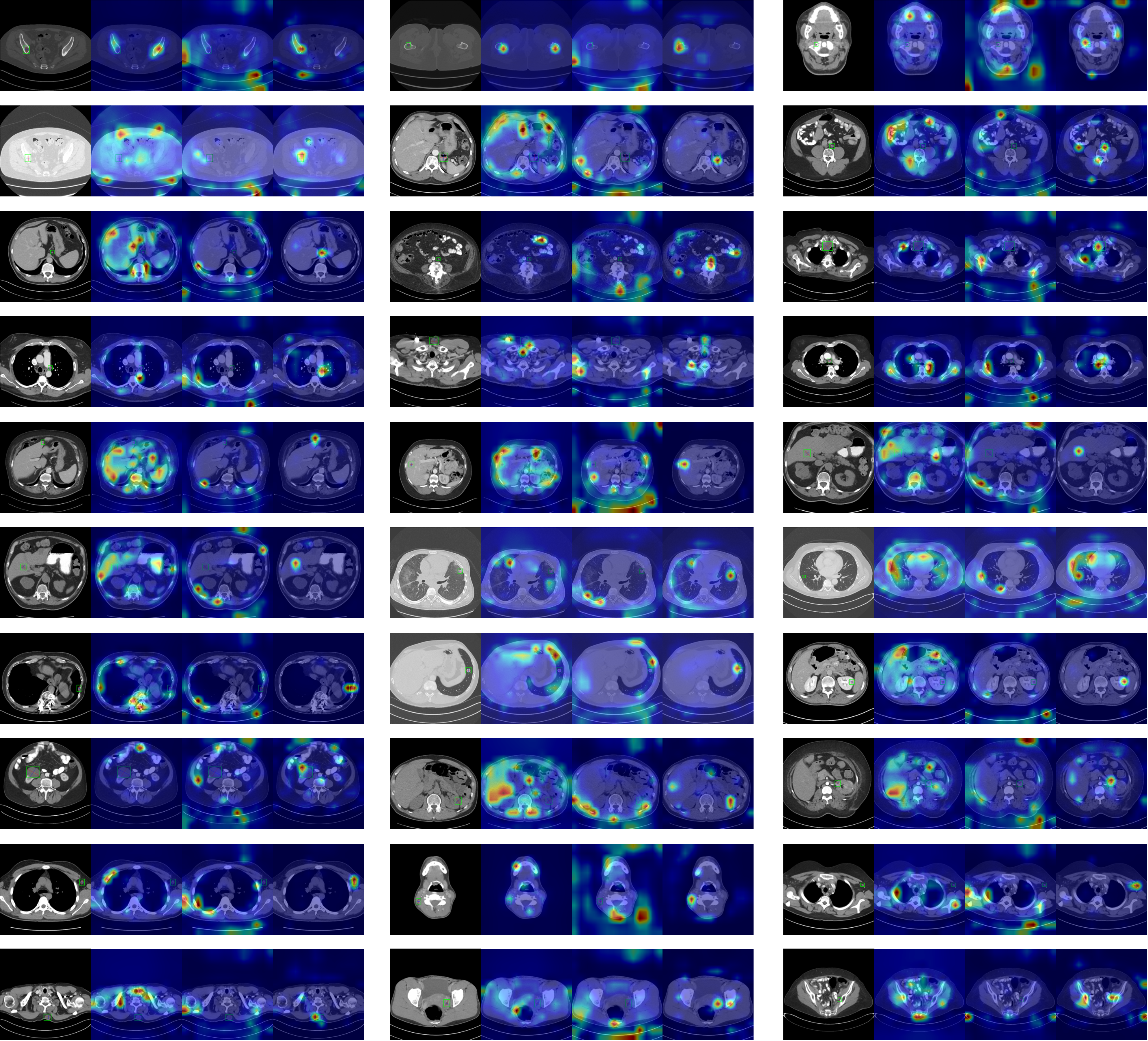}
   \caption{Results of attention visualization of MSMAE with other algorithms in the pre-training phase. 
   We visualized the self-attention of the CLS token in the last layer of the encoder. 
   From left to right, the input image, MAE attention visualization result, AMT attention visualization result, and MSMAE attention visualization result. 
   The \textcolor{green}{green box} indicates the human tissue associated with the lesion. Please zoom in to see the details better.}
   \label{figs2}
\end{figure}

\begin{figure}[h]
  \centering
  \includegraphics[width=1\linewidth]{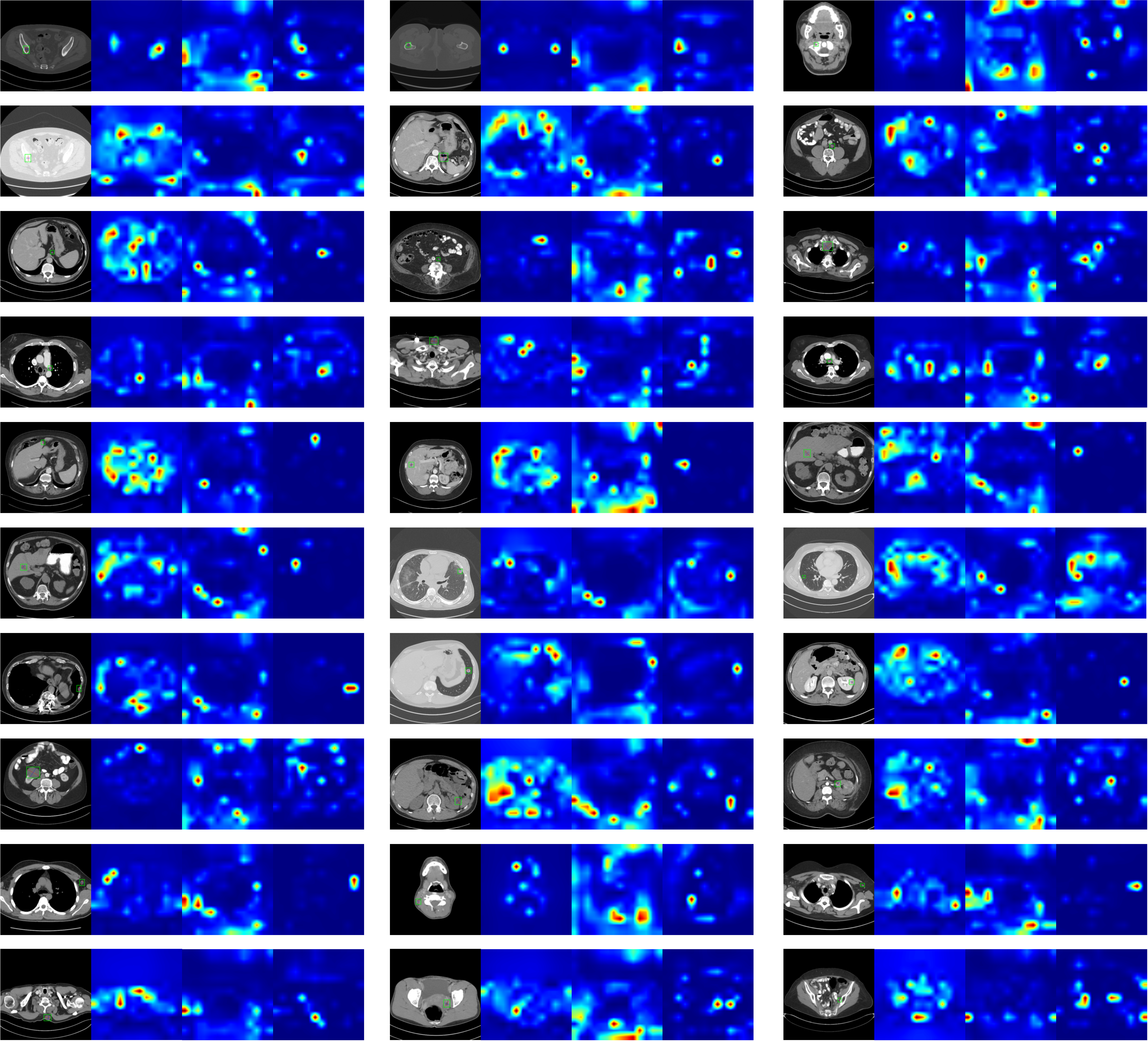}
   \caption{Comparison of the masking weights of different algorithms in the pre-training phase. 
   From left to right, the input image, MAE masking weight, AMT masking weight, and MSMAE masking weight are shown. 
   The \textcolor{green}{green box} indicates the human tissue associated with the lesion. Please zoom in to see the details better.}
   \label{figs3}
\end{figure}

\begin{figure}[h]
  \centering
  \includegraphics[width=1\linewidth]{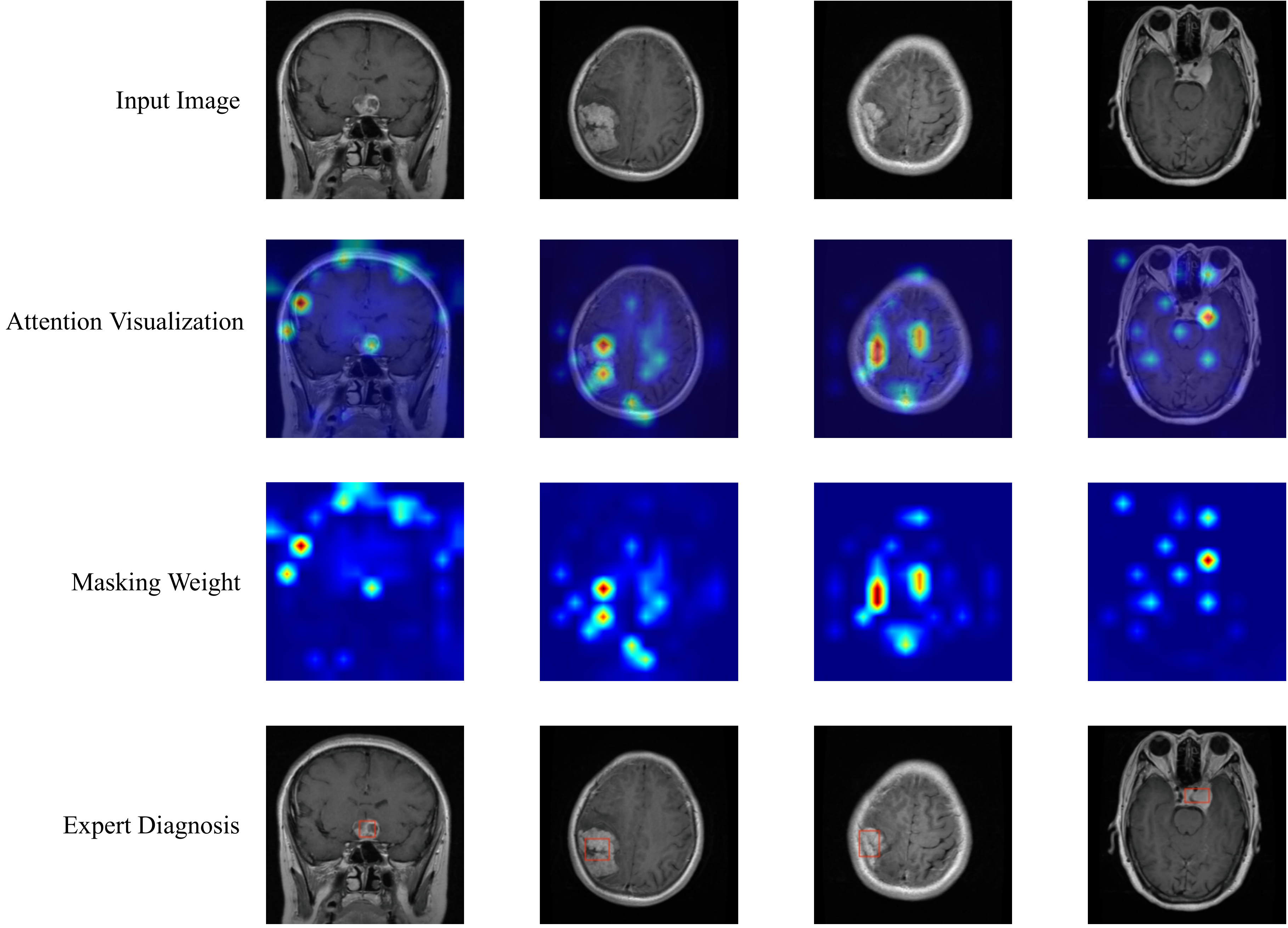}
   \caption{Attention visualization results and masking weights of MSMAE on the BTMD dataset labeled samples in the pre-training phase. 
   The \textcolor{red}{red box} indicates the human tissue associated with the lesion as marked by the expert. Please zoom in to see the details better.}
   \label{figs4}
\end{figure}

\begin{figure}[h]
  \centering
  \includegraphics[width=1\linewidth]{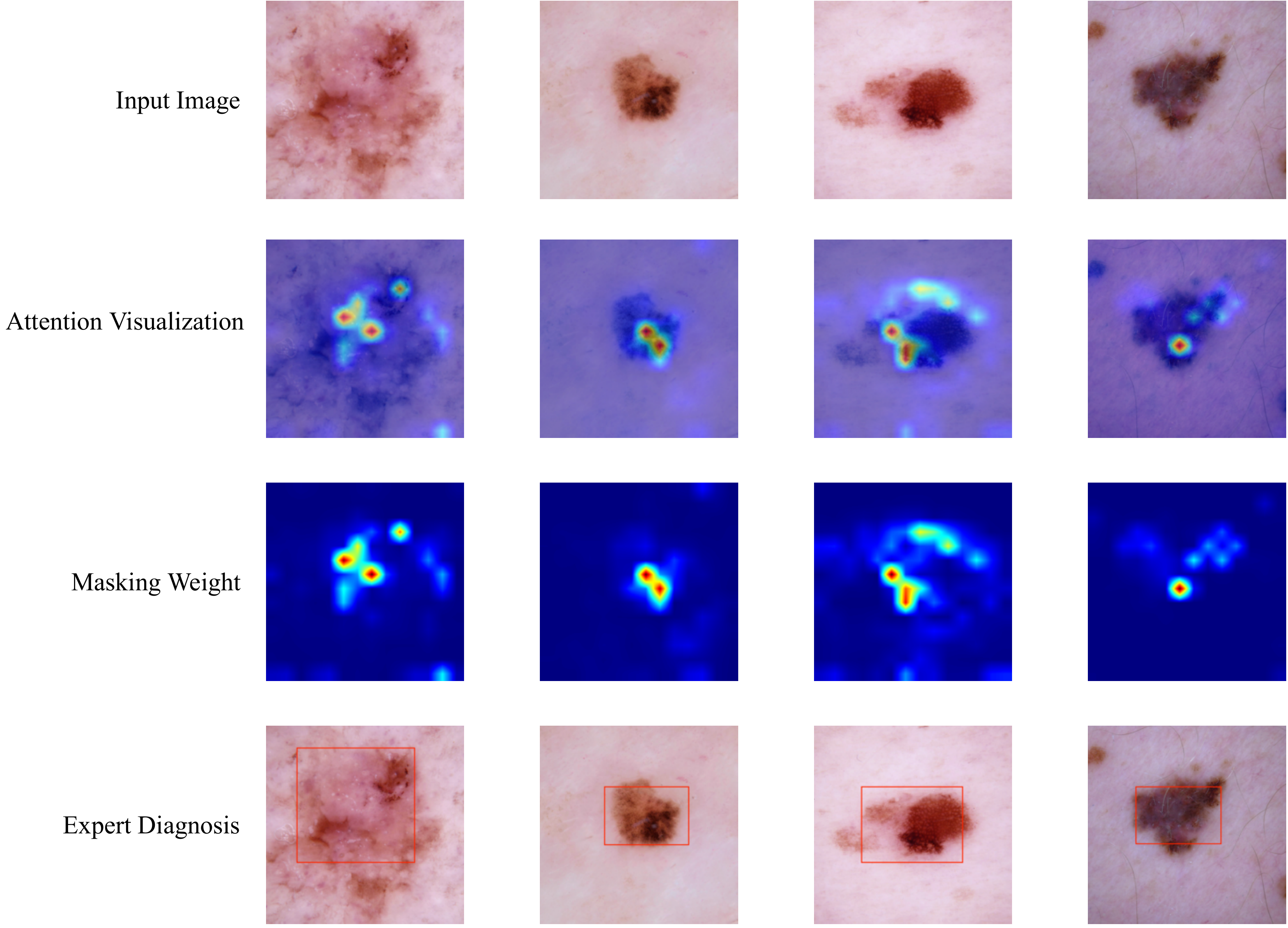}
   \caption{Attention visualization results and masking weights of MSMAE on the HAM10000 dataset labeled samples in the pre-training phase. 
   The \textcolor{red}{red box} indicates the human tissue associated with the lesion as marked by the expert. Please zoom in to see the details better.}
   \label{figs5}
\end{figure}

\begin{figure}[h]
  \centering
  \includegraphics[width=1\linewidth]{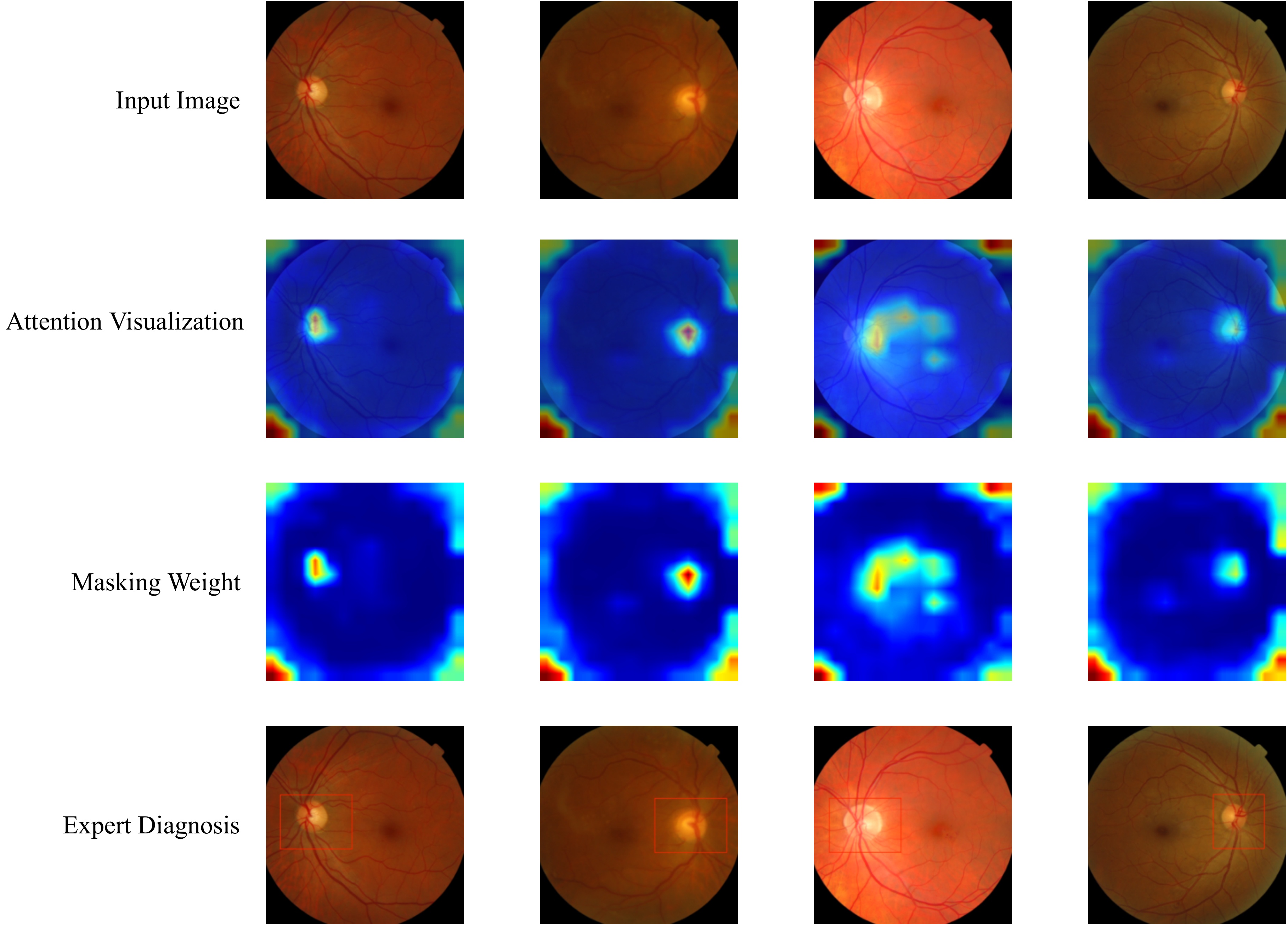}
   \caption{Attention visualization results and masking weights of MSMAE on the Messidor-2 dataset labeled samples in the pre-training phase. 
   The \textcolor{red}{red box} indicates the human tissue associated with the lesion as marked by the expert. Please zoom in to see the details better.}
   \label{figs6}
\end{figure}

In this section, we give more results on the comparison of MSMAE with other algorithms in the pre-training phase of attention visualization and masking weights in Fig. \ref{figs2} and Fig. \ref{figs3}. 
Moreover, we have invited experts to annotate some samples (see \textcolor{red}{red box} in Fig. \ref{figs4}, Fig. \ref{figs5}, and Fig. \ref{figs6}) in our medical image classification datasets. 
And we visualized the attention maps of these labeled samples and their masking weights using the pre-trained MSMAE.
The specific results are shown in Fig. \ref{figs4}, Fig. \ref{figs5}, and Fig. \ref{figs6}.

\clearpage

{
\small

\bibliographystyle{plain}
\bibliography{egbib.bib}

}

\end{document}